\documentclass[final]{clv2plain}
\pdfoutput=1

\usepackage[fleqn]{amsmath}
\usepackage{amsfonts}
\usepackage{amssymb}
\usepackage{amsbsy}
\usepackage{xspace}

\makeatletter                                                           
\newcommand{\@BIBLABEL}{\@emptybiblabel}                                
\newcommand{\@emptybiblabel}[1]{}                                       
\makeatother                                                            
\usepackage[hidelinks]{hyperref}
\usepackage{latexsym}
\usepackage{booktabs}

\usepackage[inline,shortlabels]{enumitem}
\setlist*[enumerate,1]{label=$(\arabic*)$}
\setlist*[itemize,1]{label=$\bullet$}

\usepackage[T1]{fontenc}
\usepackage[utf8]{inputenc}
\usepackage[russian,english]{babel}

\renewcommand{\vec}{\boldsymbol}   % optional

\newcommand{\valpha}{{\vec{\alpha}}}

\newcommand{\vtheta}{{\vec{\theta}}}
\newcommand{\veta}{{\vec{\eta}}}

\newcommand{\vbeta}{{\vec{\beta}}}

\newcommand{\dirac}[2]{\delta_{#1}\left(#2\right)}
\newcommand{\Discrete}{\ensuremath{\mathrm{Discrete}}}
\newcommand{\Dirichlet}{\ensuremath{\mathrm{Dirichlet}}}

\newcommand{\DR}{\ensuremath{\textrm{DR}}}

\newcommand{\wfa}{\textsuperscript{$\star$}\xspace}
\newcommand{\wfb}{\textsuperscript{$\dagger$}\xspace}
\newcommand{\wfc}{\textsuperscript{$\ddagger$}\xspace}
\newcommand{\wfd}{\textsuperscript{$\star\star$}\xspace}

\def\eqref#1{Eq.~\ref{eqn:#1}}

\hypersetup{pdfauthor={Chandler May, Ryan Cotterell, and Benjamin Van Durme},pdftitle={An Analysis of Lemmatization on Topic Models of Morphologically Rich Language}}
\runningtitle{An Analysis of Lemmatization on Topic Models of Morphologically Rich Language}
\runningauthor{Chandler May, Ryan Cotterell, and Benjamin Van Durme}

\begin{document}

\pagestyle{plain}

\title{An Analysis of Lemmatization on Topic Models of Morphologically Rich Language}

\author{Chandler May}
\affil{Johns Hopkins University}

\author{Ryan Cotterell}
\affil{Johns Hopkins University}

\author{Benjamin Van Durme}
\affil{Johns Hopkins University}

\maketitle

\begin{abstract}
Topic models are typically represented by top-$m$ word lists for
human interpretation.  The corpus is often pre-processed with
lemmatization (or stemming) so that those representations are not
undermined by a proliferation of words with similar meanings, but
there is little public work on the effects of that pre-processing.
Recent work studied the effect of stemming on topic models of English
texts and found no supporting evidence for the practice.  We study
the effect of lemmatization on topic models of Russian Wikipedia
articles, finding in one configuration that it significantly improves
interpretability according to a word intrusion metric.  We conclude
that lemmatization may benefit topic models on morphologically rich
languages, but that further investigation is needed.
\end{abstract}

\section{Introduction}\label{sec:introduction}

Topic modeling is a standard tool for unsupervised analysis of large
text corpora. At the core, almost all topic models pick up on
co-occurrence signals between different words in the corpus, that is,
words that occur often in the same sentence are likely to belong to
the same latent topic. In languages that exhibit rich inflectional
morphology, the signal becomes weaker given the proliferation of
unique tokens.  While lemmatization (or stemming)
is often used to preempt this problem, its effects on a topic model are
generally assumed, not measured.  In this study we establish the first
measurements of the effect of token-based lemmatization on topic models
on a corpus of morphologically rich language.

Syntactic information is not generally considered to exert a strong
force on the thematic nature of a document.  Indeed, for this reason
topic models often make a bag-of-words assumption, discarding the order
of words within a document.  In morphologically rich languages,
however, syntactic information is often encoded in the word form
itself.  This kind of syntactic information is a nuisance variable
in topic modeling and is prone to polluting a topic representation
learned from data~\cite{boydgraber2014}.
For example, consider the Russian name
{\em Putin}; in English, we have a single type that represents {\em
Putin} in all syntactic contexts, whereas in Russian
{\selectlanguage{russian} Путин} appears with various inflections,
e.g., {\selectlanguage{russian}Путина},
{\selectlanguage{russian}Путину}, {\selectlanguage{russian}Путине},
and {\selectlanguage{russian}Путином}. Which form of the name one uses
is fully dependent on the syntactic structure of the sentence. Compare
the utterances {{\selectlanguage{russian}мы говорим о Путине} ({\em we
    are speaking about Putin}) and {{\selectlanguage{russian}мы
      говорим Путину} ({\em we are speaking to Putin})}: both sentences
  are thematically centered on Putin, but two different word forms
  are employed.
English stop words like prepositions often end up as inflectional
suffixes in Russian, so lemmatization on Russian performs some
of the text normalization that stop word filtering performs on English.
Topic models are generally sensitive to stop
words in English~\cite{wallach2009,blei2010,eisenstein2011}, hence we
expect them to be sensitive to morphological variation in languages
like Russian.

In this study, in a corpus of Russian Wikipedia pages, we show that
\begin{itemize}
    \item truncated documents, imitating the sparsity seen in social
        media, reduce interpretability (as measured by a word intrusion
        evaluation);
    \item if lemmatization is applied, filtering the vocabulary yields
        more interpretable topics than adding an informative (but
        fixed) prior; and
    \item overall, interpretability is best when the corpus consists
        of untruncated documents, the vocabulary is filtered, and
        lemmatization is applied.
\end{itemize}
Finally, we compare our approach and findings to a recent,
comprehensive study of stemming topic models on English
corpora~\cite{schofield2016} and offer suggestions for future work.

\section{Morphology and Lemmatization}\label{sec:inflectional}

Morphology concerns itself with the internal structure of individual
words.  Specifically, we focus on {\em inflectional morphology}, word
internal structure that marks syntactically relevant linguistic
properties, e.g., person, number, case and gender, on the word form
itself. While inflectional morphology is minimal in English and
virtually non-existent in Chinese, it occupies a prominent position in
the grammars of many other languages, like Russian. In fact, Russian will often
express relations marked in English with prepositions simply through
the addition of a suffix, often reducing the number of words in a
given sentence. The collection of inflections of the same stem is preferred to as a
{\em paradigm}.  The Russian noun, for example, forms a paradigm with 12
forms.  See the sample paradigm in Table~\ref{tab:paradigm} for an
example.\footnote{Note that Table~\ref{tab:paradigm} contains several
  entries that are identical, e.g., the singular genitive is the same
  as the singular accusative. This is a common phenomenon known as
  syncretism \cite{baerman2005syntax}, but it is not universal over all nouns---plenty of other
  Russian nouns {\em do} make the distinction between
  genitive and accusative in the singular.} The Russian verb is even more expressive with more
than 30 unique forms \cite{wade2010comprehensive}.

In natural language processing, large paradigms imply an increased token to type
ratio, greatly increasing the number of unknown words. One method to
combat this issue is to {\em lemmatize} the sentence.  A lemmatizer maps each
inflection (an element of the paradigm) to a canonical form known as
the lemma, which is typically the form found in dictionaries written
in the target language.
In this work, we employ the TreeTagger
lemmatizer \cite{schmid1994probabilistic}.\footnote{
   \url{http://www.cis.uni-muenchen.de/~schmid/tools/TreeTagger/}
}
The parameters were estimated using the Russian corpus described in
\namecite{sharov2011proper}.

\begin{table}
    \centering
  \begin{tabular}{l | l l }
    & {\bf Singular} & {\bf Plural} \\ \hline
    {\bf Nominative} &  {\selectlanguage{russian}пес} ({\em pyos}) & {\selectlanguage{russian}псы}    ({\em psy})   \\
    {\bf Genitive} &  {\selectlanguage{russian}пса} ({\em psa}) & {\selectlanguage{russian}псов}    ({\em psov})  \\
    {\bf Accusative} &  {\selectlanguage{russian}пса} ({\em psa}) & {\selectlanguage{russian}псов}    ({\em psov})  \\
    {\bf Dative} &  {\selectlanguage{russian}псу} ({\em psu}) & {\selectlanguage{russian}псам}    ({\em psam})  \\
    {\bf Locative} &  {\selectlanguage{russian}псе} ({\em psye}) & {\selectlanguage{russian}псах}   ({\em psax})  \\
    {\bf Instrumental} &  {\selectlanguage{russian}псом} ({\em psom}) & {\selectlanguage{russian}псами}  ({\em psami}) \\
  \end{tabular}
  \caption{A inflectional paradigm for the Russian word
    {\selectlanguage{russian}пес} ({\em pyos}), meaning ``dog''.  Each
    of the 12 different entries in the table occurs in a distinct
    syntactic context. A lemmatizer canonicalizes these forms to
    single form, which is the nominative singular in, reducing the sparsity present in the corpus.}
    \label{tab:paradigm}
\end{table}

\section{Related Work}\label{sec:related-work}

To measure the effect of lemmatization on topic models, we must first
define the term ``topic models.''  In this study, for comparability with other
work, we restrict our attention to latent Dirichlet allocation
(LDA)~\cite{blei2003}, the canonical Bayesian graphical topic model.
We want to measure the performance of a topic model by its
interpretability, as topic models are best suited to discovering
human-interpretable decompositions of the data~\cite{may2015}.
We note there are more modern but less widely-used topic models than
LDA, such as the sparse additive generative
(SAGE) topic model, which explicitly models the background word
distribution and encourages sparse topics~\cite{eisenstein2011}, or the
nested hierarchical Dirichlet process (nHDP) topic model, which
represents topics in a hierarchy and automatically infers its effective
size~\cite{paisley2015}.  These models may be more interpretable by
some measures but are less widely used and accessible.
Separately, the infinite-vocabulary LDA model has a
prior similar to an $n$-gram model~\cite{zhai2013}, which could be
viewed as loosely encoding beliefs of a concatenative morphology, but
the effect of that prior has not been analyzed in isolation.
We seek to measure the impact of
lemmatization on a topic model and would like our results to be
applicable to research and industry, so we leave these
alternative topic models as considerations for future work.

Though stemming and lemmatization have long
been applied in topic modeling
studies~\cite{deerwester1990,hofmann1999,mei2007,nallapati2008,lin2009},
their effect on a topic model was publicly investigated only recently, in a
comparison of rule-based and context-based stemmers in LDA topic models on four
English corpora~\cite{schofield2016}.
Overall, stemming was found to reduce model fit, negligibly
affect topic coherence, and negligibly or
negatively affect model consistency across random initializations.
In light of these results, \namecite{schofield2016}
recommended refraining from stemming the corpus as a pre-processing
step and instead stemming the top-$m$ word lists as a post-processing
step, as needed.
Our analysis is more narrow, and complementary: we measure the
interpretability of topics using a word intrusion metric on a
single, distinct corpus of a morphologically richer language; we
evaluate a single, distinct lemmatizer; we also use fixed
hyper-parameters, stochastic variational inference rather than Gibbs
sampling, and 100 rather than 10, 50, or 200 topics.  Thus, while the
difference in morphological variation is an intuitive explanation of
our different conclusions, it is by no means the only explanation.

Though its analysis is nascent in the context of topic
modeling, morphology has been actively investigated in the context of
word embeddings.  Topic proportions parallel continuous embeddings:
both are real-valued representations of lexical semantic
information. Most notably, \namecite{BianGL14} learned
embeddings for individual morphemes jointly within the standard {\sc word2vec}
model \cite{mikolov2013distributed} and \namecite{SoricutO15} used the embeddings
themselves to induce morphological analyzers. Character-level embedding approaches
have also been explored with the express aim of capturing morphology \cite{santos2014learning,LingDBTFAML15}.

\begin{table*}
    \centering
    \begin{tabular}{l|l}
        view & topic \\\hline

        lem & {\selectlanguage{russian}деревня\wfa сельский поселение пункт сельсовет} \\
        non & {\selectlanguage{russian}деревня\wfa деревни\wfa деревне\wfa жителей волости} \\\hline

        lem & {\selectlanguage{russian}клетка лечение\wfa заболевание\wfb препарат действие} \\
        non & {\selectlanguage{russian}лечения\wfa течение лечение\wfa крови заболевания\wfb} \\\hline

        lem & {\selectlanguage{russian}японский\wfa япония\wfb корея префектура смотреть} \\
        non & {\selectlanguage{russian}считается японии\wfb японский\wfa посёлок японской\wfa} \\\hline

        lem & {\selectlanguage{russian}художник\wfa искусство\wfb художественный\wfa картина\wfc выставка\wfd} \\
        non & {\selectlanguage{russian}искусства\wfb музея картины\wfc выставки\wfd выставка\wfd} \\
    \end{tabular}
    \caption{Manually-aligned topic pairs: the first topic in each pair
        is from the lemmatized model, the second pair is a semantically
        similar topic in the non-lemmatized model.  Within each pair,
        each of the symbols \wfa, \wfb, \wfc, and \wfd (separately)
        denotes word forms of a shared lemma.
        The lemmatized topic representations are more
        diverse than those of the non-lemmatized topic representations.
        For example, the non-lemmatized version of the first topic
        contains three inflections of the Russian word
        {\selectlanguage{russian}деревня} ({\em village})---successive
        inflectional forms add little or no information to the topic.
    }
    \label{tab:topics}
\end{table*}

\section{Experiments}\label{sec:experiments}

For some pre-specified
number of topics $K$ and Dirichlet concentration hyperparameters
$\veta$ and $\valpha$, the LDA topic model represents a vocabulary as a
set of $K$ independent and identically distributed (i.i.d.) topics $\vbeta_k$, represents each document as a
an i.i.d.\ mixture over those topics (with mixture weights
$\vtheta_d$), and specifies that each token in a document is
generated by sampling a word type from the document's topic mixture:
\begin{align*}
    \vbeta_k  & \sim \Dirichlet\left(\veta\right) \\
    \vtheta_d & \sim \Dirichlet\left(\valpha\right) \\
    z_{d,n}              & \sim \Discrete\left(\vtheta_d\right) \\
    w_{d,n}              & \sim \Discrete\left(\vbeta_{z_{d,n}}\right)
\end{align*}

Meaningful evaluation of topic models is notoriously
difficult and has received considerable attention in the
literature~\cite{chang2009,wallach2009a,newman2010,mimno2011,lau2014}.
In general we desire an evaluation metric that correlates with a
human's ability to use the model to explore or filter a large dataset,
hence, the interpretability of the model.  In this study, we moreover
require an evaluation metric that is comparable across different views
of the same corpus.

With those concerns in mind we choose a \emph{word intrusion}
evaluation:
a human expert is shown one topic at a time, represented
by its top $m$ words (for some small number $m$) in random order, as
well as an additional word (called the \emph{intruder}) randomly placed
among the $m$ topic words~\cite{chang2009}.
The intruder is randomly selected from the set of high-probability
words from other topics in the model.
The expert is tasked with identifying the intruder in each list of
$m + 1$ words.
As in prior work~\cite{chang2009}, we instruct the expert to ignore
syntactic and morphological patterns.

If the model is interpretable, the $m$ words from a topic will be
internally coherent whereas the intruder word is likely to stand out.
Thus a model's interpretability can be quantified by the fraction
of topics for which the expert correctly identifies the intruder.  We
call this value the \emph{detection rate}:
\begin{equation*}
    \DR = \frac{1}{K} \sum_{k=1}^K \dirac{i_k}{\omega_k}
\end{equation*}
where $K$ is the number of topics in the model, $i_k$ is the index
of the intruder in the randomized word list generated from topic $k$,
and $\omega_k$ is the index of the word the expert identified as the
intruder.  We note this is just the mean (over topics) of the
\emph{model precision} metric from prior work~\cite{chang2009}
when one expert is used instead of several non-experts.

Our corpus consists of Russian Wikipedia articles from the dump
released on 11/02/2015.\footnote{The Wikipedia dump is from November 11, 2015.}
We stripped the XML portion of the formatting and then ran the
lemmatizer described in Section~\ref{sec:inflectional}.  When the
lemmatizer does not recognize a word, we back off to the word form
itself.\footnote{
    11\% of the 378 million tokens in the raw corpus were
    unrecognized by the lemmatizer.
}

We consider two preprocessing schemes to account for stop words and
other high-frequency terms in the corpus.  First, we compute the
vocabulary as the top 10,000 words by document frequency,\footnote{
    Due to minor implementation concerns the lemmatized and
    non-lemmatized vocabularies consist of the top 9387 and 9531 words
    (respectively) by document frequency.
}
separately for the lemmatized and non-lemmatized data, and
specify an asymmetric prior on each document's topic proportions
$\vtheta$.
We refer to this preprocessing scheme
as the \emph{unfiltered-asymmetric} setting.  The second modeling scheme we
consider uses a vocabulary with high-frequency words filtered out and a
uniform prior on the document-wise topic proportions.
(We refer to this setting as \emph{filtered-symmetric}.)
Specifically, a 10,000 word vocabulary is formed from the
lemmatized data by removing the top 100 words by document frequency
over the corpus and taking the next 10,000.  To determine the
non-lemmatized vocabulary, we map the filtered lemmatized
vocabulary onto all word forms that produce one of those lemmas in
the data.  Finally, observing that some of the uninformative
high-frequency words reappear in this projection, we remove any
of the top 100 words from the lemmatized and non-lemmatized corpora
from this list, producing a non-lemmatized vocabulary of 72,641 words.
While the large size of this vocabulary slows learning,
we do not believe it impacts the results negatively;
our priority is retaining the information captured by the lemmatized
vocabulary to provide a fair comparison.

In addition to exploring different choices of vocabulary, we also
consider truncating the documents to their first 50 tokens.\footnote{
    As the vocabulary does not contain rare words, the number of
    tokens per document seen by the model is less than 50.
}
This augmentation simulates data sparsity by reducing the amount of
content-bearing signal in each document, so we might expect the
truncated documents to more greatly benefit from lemmatization (which
can be cast as a dimensionality reduction method).

We learn LDA by stochastic variational
inference~\cite{hoffman2013}, initializing the models randomly and
using fixed priors.\footnote{
    In preliminary experiments Gibbs
    sampling with hyper-parameter optimization did not improve
    interpretability.
}
We specify $K = 100$ topics to all models.
Uniform priors with $\eta_v = 0.1$ and
$\alpha_k = 5 / K$ were given to
filtered-symmetric models; non-uniform priors with
$\eta_v = 0.1$, $\alpha_1 = 5$, and $\alpha_k = 5 / (K-1)$
for $k > 1$
were given to unfiltered-asymmetric models.
The local hyperparameters $\valpha$ are informed by mean
document word usage and document length; in particular, we
believe approximately 50\% of the word tokens in the corpus are
uninformative.

The detection rate for all four configurations (filtered-symmetric or unfiltered-asymmetric
vocabulary and full-length or truncated documents), and the
p-values for one-sided detection rate differences (testing our
hypothesis that the lemmatized models yield higher detection rates than
the non-lemmatized models), are reported in
Table~\ref{tab:detection-rate}.  Word intrusion performance benefits
significantly from lemmatization on a filtered vocabulary and a
symmetric prior.
Truncated documents exhibit lower
performance overall and are helped less by lemmatization (posing
challenges for social media applications).
Further, we observe differences
between use of an asymmetric prior on an unfiltered vocabulary and
use of a symmetric prior on a vocabulary with stop words filtered out.

\begin{table}
  \centering
    \begin{tabular}{rrr|rr|r}
              &              &                & \multicolumn{2}{|c|}{$\DR$} &       p-val   \\\hline
        vocab & prior        & docs           & non         & lem         &         $\Delta$ \\\hline
        unfilt & sym         & full           &          0.54 &          0.52 &          0.61 \\
        \textbf{filt} & \textbf{asym} & \textbf{full} & \textbf{0.50} & \textbf{0.65} & \textbf{0.02} \\
        unfilt & sym         & trunc          &          0.37 &          0.37 &          0.50 \\
        filt & asym          & trunc          &          0.43 &          0.47 &          0.28 \\
    \end{tabular}
    \caption{Detection rate for the non-lemmatized (non) and
        lemmatized (lem) models
        and p-values for the one-sided detection rate difference tests.
        (filt and unfilt indicate whether or not the vocabulary is
        filtered; sym and asym indicate whether the prior is symmetric,
        trunc and full indicate whether the documents are truncated.)
        The detection rate benefits significantly from lemmatization on
        a filtered vocabulary (highlighted in bold).}
    \label{tab:detection-rate}
\end{table}

We find that topics from the unfiltered-asymmetric models often contain
stop words despite the first topic receiving half of the prior
probability mass.  Indeed, many topics consist primarily of stop
words, such as the topic {\selectlanguage{russian}и в при с у}.
Hand-aligned topics from the filtered-symmetric models learned on
full-length documents are shown in Table~\ref{tab:topics}.
There is significant redundancy (multiple inflected word forms of the
same lemma) in the top five words of the non-lemmatized topics; on the
other hand, the diversity of words in the lemmatized topics lends
to human interpretation.

\section{Conclusion}\label{sec:conclusion}

We have measured the effect of pre-processing by lemmatization
on the interpretability of topic models in a morphologically rich
language.  Unlike a prior study on English, we found empirical
justification for this intuitive but largely unexamined practice.
However, our approach is distinct, and further work is required to determine what factors
contribute to our different conclusions.  In the meantime,
we recommend measuring rather than assuming the effects of
lemmatization (or stemming) in a new topic modeling application,
and we echo the suggestion of prior work to stem during post-processing
\emph{as needed}~\cite{schofield2016}.

\begin{acknowledgments}
    We would like to thank the Johns Hopkins Human Language Technology
    Center of Excellence and the DARPA LORELEI program for providing
    support.
    The second author acknowledges support from a DAAD long-term
    research grant.
    Any opinions expressed in this work are those of the authors.
\end{acknowledgments}

\bibliographystyle{fullname}
\bibliography{russian}

\end{document}